# Wisdom of Crowds Cluster Ensemble

Hosein Alizadeh[1], Muhammad Yousefnezhad[2] and Behrouz Minaei Bidgoli[3]

**Abstract:** The *Wisdom of Crowds* is a phenomenon described in social science that suggests four criteria applicable to groups of people. It is claimed that, if these criteria are satisfied, then the aggregate decisions made by a group will often be better than those of its individual members. Inspired by this concept, we present a novel feedback framework for the cluster ensemble problem, which we call Wisdom of Crowds Cluster Ensemble (WOCCE). Although many conventional cluster ensemble methods focusing on diversity have recently been proposed, WOCCE analyzes the conditions necessary for a crowd to exhibit this collective wisdom. These include decentralization criteria for generating primary results, independence criteria for the base algorithms, and diversity criteria for the ensemble members. We suggest appropriate procedures for evaluating these measures, and propose a new measure to assess the diversity. We evaluate the performance of WOCCE against some other traditional base algorithms as well as state-of-the-art ensemble methods. The results demonstrate the efficiency of WOCCE's aggregate decision-making compared to other algorithms.

## 1. Introduction

Clustering, one of the main tasks of data mining, is used to group non-labeled data to find meaningful patterns. Generally, different models provide predictions with different accuracy rates. Thus, it would be more efficient to develop a number of models using, different data subsets, or utilizing differing conditions within the modeling methodology of choice, to achieve better results. However, selecting the *best* model is not necessarily the ideal choice because potentially valuable information may be wasted by discarding the results of less-successful models (Perrone and Cooper, 1993; Tumer and Ghosh, 1996; Baker and Ellison, 2008).

[1]Hosein Alizadeh
Department of Computer Eng., Iran University of Science and Technology, Tehran, Iran
e-mail: halizadeh@iust.ac.ir

[2]Muhammad Yousefnezhad
Computer Science and Technology Department, Nanjing University of Aeronautics and Astronautics, Nanjing, China
e-mail: myousefnezhad@nuaa.edu.cn

[3]Behrouz Minaei-Bidgoli
Department of Computer Eng., Iran University of Science and Technology, Tehran, Iran
e-mail: b_minaei@iust.ac.ir

This leads to the concept of *combining*, where the outputs (individual predictions) of several models are pooled to make a better decision (collective prediction) (Tumer and Ghosh, 1996; Baker and Ellison, 2008). Research in the *Clustering Combination* field has shown that these pooled outputs have more strength, novelty, stability, and flexibility than the results provided by individual algorithms (Strehl and Ghosh, 2002; Topchy et al., 2003; Fred and Lourenco, 2008; Ayad and Kamel, 2008).

In the classic cluster ensemble selection methods, a consensus metric is used to audit the basic results in cluster ensemble selection and use them to produce the final result. There are two problems in the traditional methods; firstly, although the final result is always in accordance with the selected metrics providing the optimized result, there might be other metrics by which a better final result can be generated. Secondly, In order to produce the final result, there is neither any information from other entities of cluster ensemble except auditing basic results and nor that any evaluation of information and errors in other entities can be presented. In order to solve the mentioned problems, this paper proposes wised clustering (WOCCE) as a viable solution. This method audits all entities of cluster ensemble and the errors in result from each entity optimized by information obtained from other entities which dramatically reduces the possibility of any errors to occur in complex data as the result.

In the social science arena, there is a corresponding research field known as the *Wisdom of Crowds*, after the book by the same name (Surowiecki, 2004), simply claiming that the Wisdom of Crowds (WOC) is the phenomenon whereby decisions made by aggregating the information of groups usually have better results than those made by any single group member. The book presents numerous case studies and anecdotes to illustrate its argument, and touches on several fields, in particular economy and psychology. Surowiecki justifies his own theory, stating that: "If you ask a large enough groups of diverse and independent people to make a prediction or estimate a probability, the average of those answers, will cancel out errors in individual estimation. Each person's guess, you might say, has two components: information and errors. Subtract the errors, and you're left with the information" (Surowiecki, 2004).

In spite of the lack of a well-defined agreement on metrics in cluster ensembles, Surowiecki suggested a clear structure for building a wise crowd. Supported by many examples from businesses, economies, societies, and nations, he argued that a wise crowd must satisfy four conditions, namely: diversity, independence, decentralization, and an aggregation mechanism. The goal of this paper is to use the WOC in order to choose a proper subset in a cluster ensemble. Whereas Surowiecki's definition of the WOC emphasizes on social problems, and the decision elements embedded in his definitions are personal opinions. This paper proposes a mapping between cluster ensemble literature and the WOC phenomenon. According to this mapping, a new WOC Cluster Ensemble (WOCCE) framework, which employs the WOC definition of well-organized crowds, is proposed. Experimental results on a number of datasets show that in comparison with similar cluster ensemble methods, WOCCE improves the final results more efficiently.

In summary, the main contributions of this paper are:

- A new framework for generating a cluster ensemble from basic (primary) clustering results with *feedback-mechanism*. WOCCE controls the quality of the ensemble using this *mechanism*.
- A new mapping between the WOC observation (an approach to social problems) and the cluster ensemble problem (one of the main fields in data mining). This allows us to apply the definitions of a wise crowd to any cluster ensemble arena.
- A new heuristic method for measuring independence according to the wise crowd definitions.
- A new diversity metric called A3, which is based on the Alizadeh–Parvin–Moshki–Minaei (APMM) criterion Alizadeh et al. (2011, 2014). A3 measures the diversity of a partition with respect to a reference set (an ensemble).

The rest of this paper is organized as follows: Section 2 reviews some relevant literature. In Section 3, we propose our new framework, and demonstrate the results of a comparison against traditional methods in Section 4. Finally, we present our conclusion in Section 5.

# 2. Literature review

## 2.1. Cluster Ensemble

In unsupervised learning methods, *cluster ensemble* has demonstrated that better final result can be achieved by combining basic results instead of choosing only the best one. This has led to the idea of an ensemble in machine learning, where the component models (also known as members) are redundant in that each provides a solution to the same task, even though this solution may be obtained by different means (Grofman and Owen, 1996; Baker and Ellison, 2008).

Generally, a cluster ensemble has two important steps (Jain et al., 1999; Strehl and Ghosh, 2002):

1- Generating different results from primary clustering methods using different algorithms and changing the number of partitions. This step is called *generating diversity or variety*.

2- Aggregating mechanisms for combining primary results and generating the final ensemble. This step is performed by consensus functions (aggregating algorithms).

It is clear that an ensemble with a set of identical models cannot provide any advantages. Thus, the aim is to combine models that predict different outcomes, and there are four parameters -dataset, clustering algorithms, evaluation metrics, and combine methods- that can be changed to achieve this goal. A set of models can be created from two approaches: choice of data representation, and choice of clustering algorithms or algorithmic parameters. There are many consensus functions in the cluster ensemble for combining the basic results. While some of them use graph partitioners (Strehl and Ghosh 2002; Fern and Brodley 2004) or cumulative voting (Tumer et al. 2008; Ayad and Kamel 2008), others are based on co-association among base partitions (Greene et al. 2004; Fred and Jain 2005).

Halkidi et al (2001) proposed the compactness and the separation to measure the quality of clustering. Strehl and Ghosh (2002) proposed the Mutual Information (MI) metric for measuring the consistency of data partitions; Fred and Jain (2005) proposed Normalized Mutual Information (NMI), which is independent of cluster size. This metric can be used to evaluate clusters and the partition in many applications. For example, while Zhong and Ghosh (2005) used NMI for evaluating clusters in document clustering, (Kandylas et al., 2008) used it for community knowledge analysis and (Long et al, 2010) used it for evaluating graph clustering. Hadjitodorov et al (2006) proposed a selective strategy which is based on diversity. Zhou and Tan (2006) proposed an algorithm which is based on selective voting. Yi et al (2009) used resampling-based selective clustering ensembles. Fern and Lin (2008) developed a method that effectively uses a selection of the basic partitions to participate in the ensemble, and consequently in the final decision. They also employed the Sum NMI and Pairwise NMI as quality and diversity metrics, respectively, between partitions. Azimi and Fern (2009) proposed adaptive cluster ensemble selection. Limin et al (2012) used compactness and separation for choosing the reference partition in the cluster ensemble selection. They also used new diversity and quality metrics as a selective strategy. Jia et al (2012) used SIM for diversity measurement. SIM is calculated based on the NMI. Alizadeh et al. (2011, 2012 and 2014) have explored the disadvantages of NMI as a symmetric criterion. They used the APMM and MAX metrics to measure diversity and stability, respectively, and suggested a new method for building a co-association matrix from a subset of base cluster results. This paper uses A3 for diversity measurement which works base on the APMM measure. Additionally, we use the co-association matrix construction scheme of Alizadeh et al. (2011 and 2014). A3 and the co-association matrix are discussed in detail in Sections 3.1 and 3.4, respectively.

**2.2. The Wisdom of Crowds**

*The Wisdom of Crowds* (Surowiecki, 2004) presents numerous case studies, primarily in economics and psychology, to illustrate how the prediction performance of a crowd is better than that of its individual members. The book relates to diverse collections of independent individuals, rather than crowd psychology as traditionally understood. Its central thesis, that a diverse collection of individuals making independent decisions is likely to make certain types of decisions and predictions better than individuals or even experts, draws many parallels with statistical sampling, but there is little overt discussion of statistics in the book. Mackey (Mackey 1841) mentions that not every crowd is wise. These key criteria separate wise crowds from irrational ones (Surowiecki, 2004):

***Diversity of opinion:*** Each person has private information, even if it is only an eccentric interpretation of the known facts.

***Independence:*** People's opinions are not determined by the opinions of those around them.

***Decentralization:*** People are able to specialize and draw on local knowledge.

***Aggregation:*** Some mechanism exists for turning private judgments into a collective decision.

It is important to note that, under some conditions, the cooperation of the crowd will fail because of the consciousness of its members about each other's opinion. This will lead them to conform rather than think differently. Although each member of the crowd may attain greater knowledge and intelligence by this effect, definitely the whole crowd as a whole will become trapped into less unwise (Mackey, 1841; Page, 2007; Hadzikadic and Sun, 2010).

In recent years, the WOC has been used in the field of machine learning. Steyvers et al. (2009) used WOC for recollecting order information, and Miller et al. (2009) proposed an approach to the rank ordering problem. Welinder et al. (2010) used a multidimensional WOC method to estimate the underlying value (e.g., the class) of each image from (noisy) annotations provided by multiple annotators. WOC has also been applied to underwater mine classification with imperfect labels (Williams, 2010) and minimum spanning tree problems (Yi et al., 2010). Finally, Baker and Ellison (2008) proposed a method for using the WOC in ensembles and modules in environmental modeling.

## 3. The WOCCE approach

Surowiecki (2004) has outlined the conditions that are necessary for the crowd to be wise: *diversity, independence, and a particular kind of decentralization*. To map the WOC to a cluster ensemble, we should restate the wise crowd requirements for the corresponding field. This section discusses these preconditions in detail for the area of clustering. It seems that the best matching between individuals and their opinions in WOC is base clustering algorithms and partitions, respectively, in the context of cluster ensembles. The goal of WOCCE, as illustrated in Figure (1), is to construct a wise crowd in the primary partition via a recursive procedure.

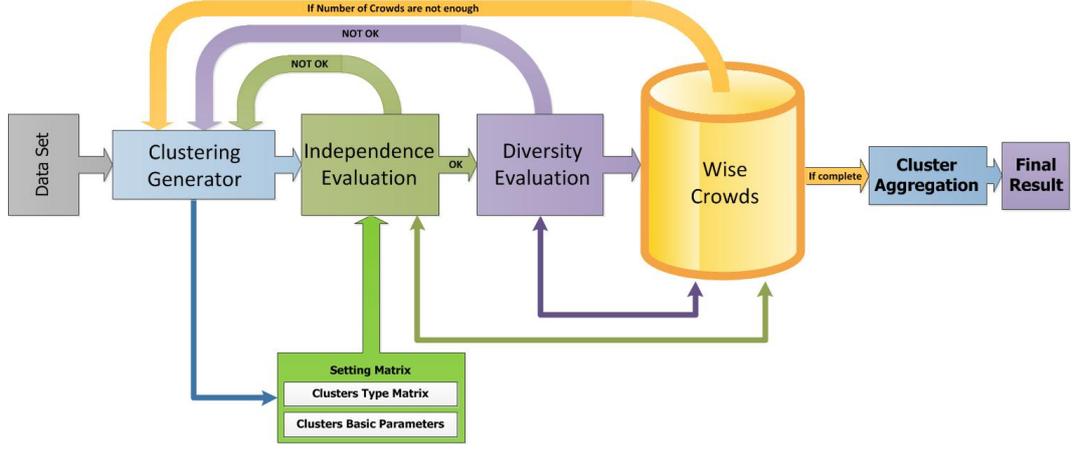

**Fig. 1**.The WOCCE framework

### 3.1. Diversity of Opinions

To define the diversity of opinion in cluster ensembles, which utilize base partitions, we use the term *diversity of base partitions*. According to this assumption, and Surowiecki's definition of diversity of opinion, it should be rephrased as:

*If the result of a base clustering algorithm has less similarity value than a defined threshold in comparison with other partitions existing in the ensemble, it is eligible to be added to the ensemble.*

Similarity and repetition of specific answers can be controlled by tracing errors. This paper proposes a new method based on APMM in order to evaluate the diversity of each primary cluster algorithm. This paper uses Average APMM, or AAPMM, to calculate the diversity of a cluster, because this method decreases the time complexity compared to NMI and avoids the symmetry problem. To calculate the similarity of cluster *C* with respect to a set of partitions P in the reference set contacting *M* partitions, we use equation (1) (Alizadeh et al., 2011):

$$AAPMM(P) = \frac{1}{M}\sum_{j=1}^{M} APMM\left(C, P_j^{b*}\right) \qquad (1)$$

Where $P_j^{b*}$ is the corresponding derivation from the *j-th* partition in the reference set. *APMM(C, P)* is the similarity between cluster *C* and a specific partition, and is given by (Alizadeh et al., 2011):

$$APMM(C,P) = \frac{-2n_c \log\left(\frac{n}{n_c}\right)}{n_c \log\left(\frac{n_c}{n}\right) + \sum_{i=1}^{kp} n_i^p \log\left(\frac{n_i^p}{n}\right)} \qquad (2)$$

In equation (2), $n_c$, $n_i^p$, and $n$ are the size of cluster *C*, the size of the *i*-th cluster of partition *p*, and the number of samples available in the partition including *C*, respectively. $k_p$ is the number of clusters in partition *P*. In order to measure the similarity of a whole partition, this paper proposes averaging AAPMM over all of the clusters that exist in a specific partition. We call this average measure *A3*. In other words, A3 is a weighted average of the AAPMMs of one partition's clusters:

$$A3(P) = \frac{1}{n} \sum_{i=1}^{k} n_i \times AAPMM(C_i) \qquad (3)$$

In equation (3), $C_i$ is the *i-th* cluster in partition *p*, and $C_i$ has size $n_i$. *n* is the number of members in partition *p* and *k* is the number of clusters in the partition. A3 measures information between a partition and those partitions in a reference set. In fact, it counts the repetition of clusters in the corresponding set. Therefore, A3 measures the similarity of a partition with respect to a set. As it is normalized between zero and one, we use *1 – A3* to represent the diversity:

$$Diversity(P) = 1 - A3(P) \qquad (4)$$

According to the above definitions, one of the conditions for appending a partition to the crowd (known as the diversity condition) is:

$$Diversity(P) \geq dT \qquad (5)$$

The threshold value for diversity is $0 \leq dT \leq 1$. Equation (5) means that if the diversity of a generated partition with respect to a set of partitions, which we call them the 'crowds' in this paper, satisfies the minimum threshold of *dT* (diversity threshold), it will be added to the crowd.

### 3.2. Independence of Opinions

According to Surowiecki's definition, independence means that an opinion must not be influenced by an individual or certain group. By mapping this to cluster ensembles, we have the following definition:

*The decision making mechanism of each base clustering algorithm must be different. In the case of using similar algorithms, the basic parameters that determine their final decisions must be sufficiently different.*

In other words, a new partition generated by a primary clustering algorithm is independent if and only if it satisfies the following conditions:

1) Every two partitions that are generated by different methods are independent because their algorithm's mechanisms are different.

2) Every two partitions that are generated by the same method with different basic parameters are independent.

This suggests that the independence of the results generated by a single algorithm should be investigated by checking the basic parameters. As most of the base algorithms are quite sensitive to their initial conditions, we propose a system of initialization checking to ensure that independent results are generated by each algorithm. The procedure Basic-Partition-Independence (BPI) illustrated in Figure (2) has been developed to calculate the independence of two partitions.

```
Function BPI (P1, P2) Return Result
    If (Algorithm-Type (P1) == Algorithm-Type (P2) then
        Result = 1 - Likeness (Basic-Parameter (P1), Basic-Parameter (P2))
    Else
        Result = 1
    End if
End Function
```

**Fig. 2.** Measuring the degree of independence between two clusters

In Figure (2), *P1* and *P2* are base partitions, the *Algorithm-Type* function returns the type of base algorithm that created those partitions, and the *Basic-Parameter* function returns the basic parameters of the algorithm that generated the partition (for example, the seed points of Kmeans). These values can be defined according to two factors: the nature of the problem and the type of base algorithms. This paper proposes a heuristic function (*Likeness*) for measuring a cluster's independence. In order to calculate the *Likeness*, we assume that $MAT_A$ and $MAT_B$ matrices contain the basic parameters of partitions $P_A$ and $P_B$, respectively. $LMAT_t$ is the distance (similarity) matrix of $MAT_A$ and $MAT_B$. $LMAT_0$ is a $n \times n$ matrix in which $n$ is the number of basic parameters in the algorithm, e.g. the number of clusters in Kmeans (because the basic parameters of Kmeans give a matrix of $k$ seed points). $LMAT_t$ contains the distances (we use a *Euclidean metric* to calculate distance) between each pair of observations in the mx-by-n data matrix $MAT_A$ and my-by-n data matrix $MAT_B$. Rows of $MAT_A$ and $MAT_B$ correspond to observations, columns correspond to variables. $LMAT_t$ is an mx-by-my matrix, with the (i, j) entry equal to distance between observation i in $MAT_A$ and observation j in $MAT_B$. $Sim_t$ is minimum value in $LMAT_t$ matrix. By removing the row and the column that contain $Sim_t$, we generate $LMAT_{t+1}$. This procedure of removing rows and columns should be repeated until **LMAT** reaches the size 0×0. As an example, Figure (3) shows how **LMAT**$_x$ matrices and **Sim**$_x$ values are calculated by using **MAT**$_A$ and **MAT**$_B$ matrices.

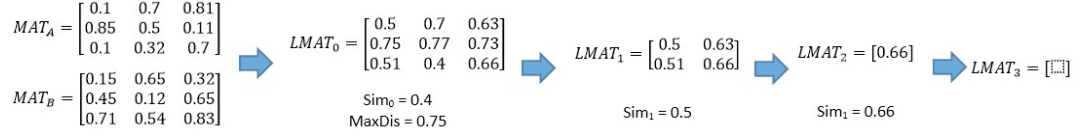

**Fig. 3.** The calculations of LMAT and Sim

Equation (6) explains the *Likeness* function. MaxDis is the maximum value in the **LMAT**$_0$ matrix in Equation (6) and Figure (3).

$$Likeness = 1 - (\frac{1}{MaxDis} \sum_{t=0}^{n} Sim_t) \qquad (6)$$

The independence of each partition is calculated as follows:

$$Independence(P) = \frac{1}{M} \sum_{i=1}^{M} BPI(P, P_i) \qquad (7)$$

Where *M* refers to the number of members in the crowd and BPI function is calculated by Figure (2) pseudo code. Thus, according to the above definitions, one of the conditions for entering the result of a clustering into the crowd is given by equation (8), which we call the independence condition and where the threshold value for Independence is $0 \leq iT \leq 1$.

$$Independence(C) \geq iT \qquad (8)$$

Based on the above definition of "Independence", this metric is not diversity metric, mainly because diversity metrics are used for evaluating the basic clustering results. Whereas independence metric controls the process of producing basic results; it is done by managing effective parameters in basic clustering algorithm. In addition, independency can calculate the probability of accuracy for similar patterns. Look at the example illustrated in Figure (4):

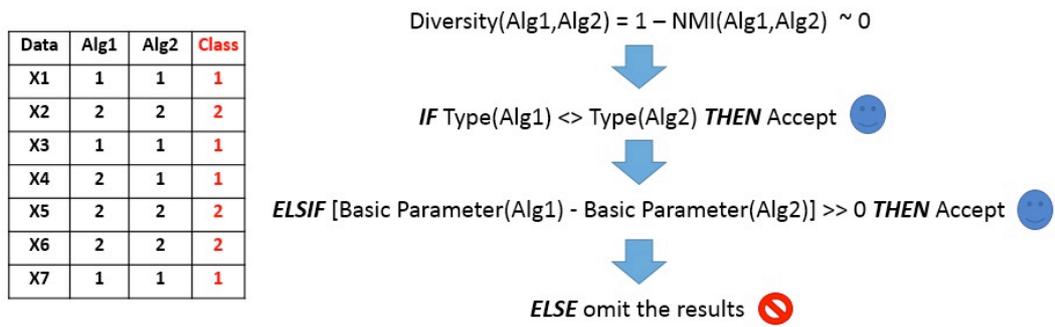

**Fig. 4.** The differences between Diversity and Independence

In this figure, Alg1 and Alg2 generate random results. As the figure shows, although the diversity is nearly zero, both algorithms' results are acceptable in comparison with the real class of dataset. Since in unsupervised learning, the real class of dataset cannot be used in the evaluation of basic algorithms' results, classic ensemble methods cannot accept the same results in ensemble committee. This paper solves this problem by proposes the Independence for predicting the probability of accuracy in primary (basic) algorithm results. According to the numeric value of the metric for two specific algorithms, if similar patterns are recognized by these two algorithms, the degree of accuracy can be identified for the pattern generated by them based on the definition of independence. Furthermore, while the independence metric analyses probability of correctness in patterns, it cannot guarantee the diversity of the final result, instead only trying to improve it. Needless to say, this paper does not intend to substitute "independence metric" with "diversity metric", rather incorporating the independence metric as a supplement to the diversity metric.

In WOCCE, the issue of algorithm's independency is considered for the first time. Independency generates repeated results in a particular redundant algorithms and ensures that the similar results are created by those algorithms with acceptable degree of independency. For this reason, the number of selected initial results in WOCCE is much smaller than the other methods. In section 4.3 of this paper, the study on the effect of Independence metric on the performance and runtime in final result will be presented.

### 3.3. Decentralization of Opinions

Surowiecki explains the necessary conditions for generating a wise crowd as follows (Surowiecki, 2004): "If you set a crowd of self-interested, independent people to <u>work in a decentralized way on the same problem, instead of trying to direct their efforts from the top down</u>, their collective solution is likely to be better than any other solution you could come up with."

According to Surowiecki's explanation of decentralization and his examples on the CIA and Linux, it can be inferred that decentralization is a quality metric. The WOC cluster should be implemented such that decentralization is established across all of its parts. According to the above, we define decentralization in a cluster ensemble as follows:

*The primary (basic) algorithm must not be influenced by any mechanism or predefined parameters; in this way, each base algorithm has a chance to reveal a 'very good result' with its own customization and local knowledge.*

In the above definition, a *very 'goodresult'* is one that has good performance, as well as enough diversity and independence to be added to the crowd. We propose two approaches; decentralization in basic results and feedback mechanism, for satisfying the notions which were defined in this paper. While decentralization in basic results satisfies decentralization conditions such as localization for basic algorithms and their input datasets, feedback mechanism tries to control decentralization conditions in ensemble components and saves the quality of results in all components.

### 3.3.1. Decentralization in basic results

This paper considers the following conditions when designing a cluster ensemble mechanism, in order to retain decentralization:

1- The number of primary algorithms participating in the crowd should be greater than one.

2- The method of entering a primary algorithm into the crowd should ensure that the final results will not be affected by its errors. In other words, the decision making of the final ensemble should not be centralized.

3- The threshold parameter *cT*, which we call the coefficient of decentralization, is a coefficient which is multiplied in the number of clusters. Every base algorithm clusters the dataset into at most *cT×k* clusters. i.e. it clusters the dataset into a number of clusters between *cT* to *cT×k*.

In the above definition, the coefficient cT is a member of natural numbers (N). This coefficient can improve accuracy in the final result when the dataset has especial complexities and basic algorithms cannot recognize the patterns in dataset. It decreases the complexities of dataset by increasing the number of clusters in basic clustering algorithms (Tan et al, 2005) and changes the complex patterns in dataset to smaller patterns which are easier to recognize by any algorithms (especially center-base-clustering algorithms). Instead of finding a complex solution for complex problems, this method turns the complex problems into smaller problems and then tries to solve them. For example, solving the Non-globular shape datasets by using center-base-clustering algorithms, such as Kmeans, can be named as one of the applications of this method. Figure (5) shows the result, achieved by applying this method on Half-Ring dataset.

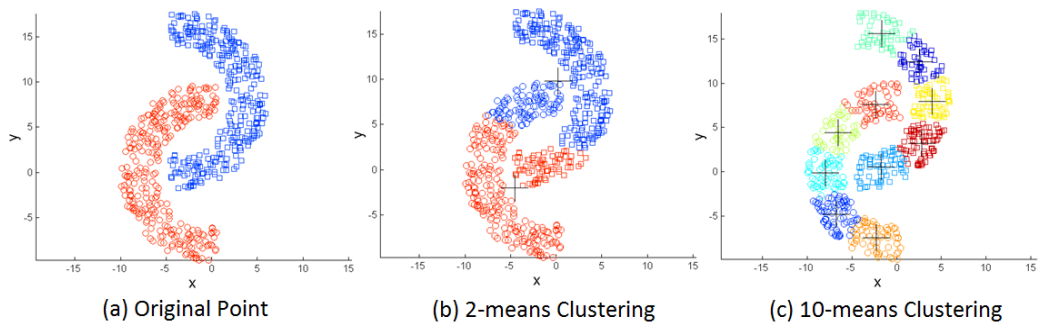

**Fig. 5.** Effect of decentralization in complex dataset

Figure (5) part (a) shows the original classes in Half-Ring dataset. Figure (5) part (b) shows the result of Kmeans algorithm when k=2 (k is equal to the original number of dataset classes). As part (b) shows, the Kmeans cannot solve this problem because it includes center-base objective function. By using this method and assuming that cT=5 (k=5x2), the Half-ring dataset complexities turn to simple patterns, as shown in part (c). These patterns can be recognized by k-means easily. In section 4.3 of this paper, the study on the effects of decentralization metric on the performance of final result will be presented.

From the above discussion, it is clear that decentralization checking should be performed during the generation of the base results. In other words, we try to satisfy the decentralization conditions in the first phase, while producing the base partitions. Therefore, unlike diversity and independence, there is no evaluation of decentralization during the assessment of the initial partitions.

**3.3.2. Feedback Mechanism**

By using feedback mechanism, our proposed method increases the number of selected results gradually. This method evaluates the basic algorithm result by using independence and diversity metrics after generating a result. If the result is accepted, it will be added to the selected set. If not, it is automatically removed. This procedure repeats itself for the next algorithm to the end. It is interesting to note that in this method, the values and their qualities do not change after selecting results, because the values are updated after every modification in each period, and that the number of selected results does not change at the end of this procedure.

Whereas in previous cluster ensemble methods, all basic clustering results had been calculated before the results were evaluated by the selected metric (e.g. NMI) and the best basic clustering results were selected by the use of thresholds. Figure (6) shows this mechanism. Although values of selected metric in our selection, in the results of basic algorithm, are maximum, this cannot guarantee that, in respect to one and others, the obtained values remain constant after being entered into our selection because the number of members in selected results changed consequently. Thus, the quality of evaluations may decrease after selecting results.

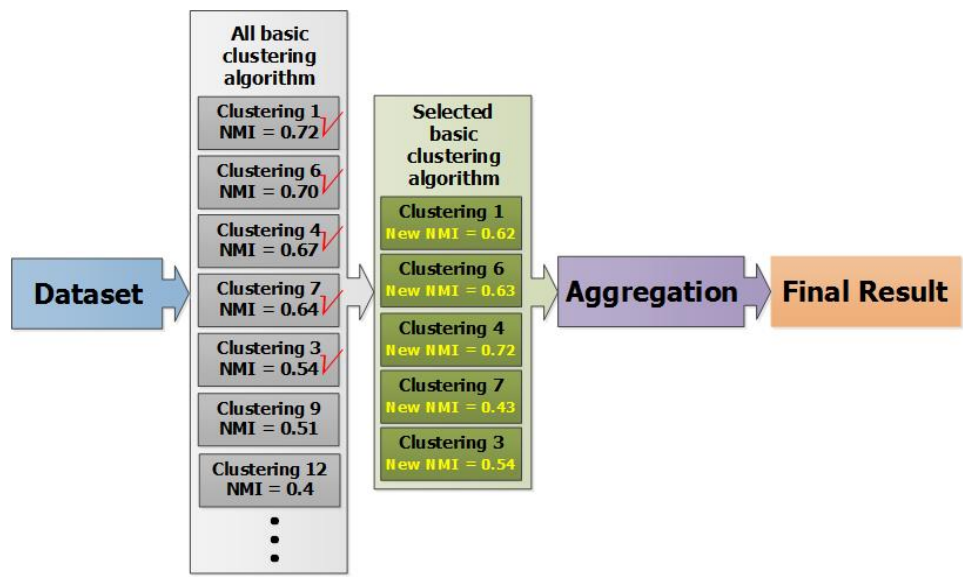

**Fig. 6.** Effect of selecting clustering in previous cluster ensemble methods on NMI values

## 3.4. Aggregation Mechanism

In this step, the opinions in the wise crowd are combined to reach a final consensus partition. In some of clustering method, the consensus partition uses a $n \times n$ co-association matrix that counts the number of groupings in the same cluster for all data points. In these methods, the primary clustering results are first used to construct the co-association matrix. The most prominent of these methods is EAC[4] (Fred and Jain, 2005). Each entry in the co-association matrix is computed as:

$$C(i,j) = \frac{n_{i,j}}{m_{i,j}} \quad (9)$$

Where $m_{i,j}$ is the number of partitions in which this pair of objects is simultaneously present and $n_{i,j}$ counts the number of clusters shared by objects with indices $i$ and $j$. WOCCE uses the co-association matrix to aggregate the results. Then employs the Average-Linkage algorithm to derive the final partition from this matrix. Figure (7) shows this process:

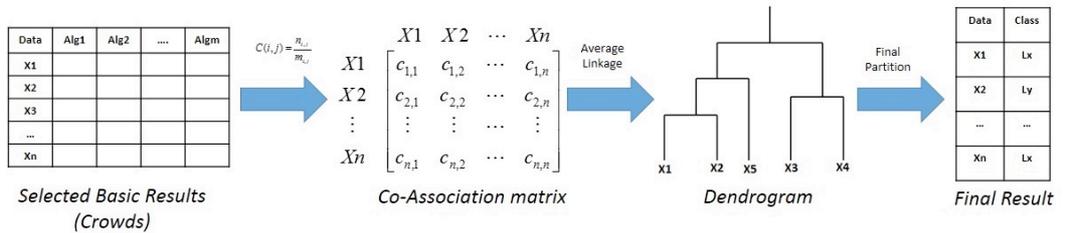

**Fig. 7.** Evidence Accumulation Clustering

In Figure (7), the selected basic results, which in this paper is called the crowds, are created by basic (primary) clustering algorithms and selected by evaluation metrics. Then, the co-association matrix is generated based on basic results. After that, the dendrogram is created by using linkage algorithms on co-association matrix (Fred and Jain, 2005; Fern and Lin, 2008; Alizadeh et al, 2011; Alizadeh et al 2012; Alizadeh et al 2014). This paper uses Average Linkage for generating dendrogram because it has high performance in comparison with other hierarchical methods in EAC (Fred and Jain, 2005; Tan et al, 2005). At last, the final result is created based on clusters' number in WOCCE.

---

[4] Evidence Accumulation Clustering

## 3.5. Summing up

In WOCCE, the process starts with an evaluation of the diversity and independence of the partitions which it is shown in Figure (1). As stated earlier, the necessary decentralization conditions are satisfied in the cluster generation phase by constructing the base partitions. Therefore, there is no component for assessing the decentralization of the generated partitions. In the WOCCE framework, only the decentralized partitions that pass both the independence and diversity filters are eligible to join the wise crowd.

In summary, the differences between these two approaches are:

1- The method of evaluating the clustering algorithm. In the WOCCE approach, the diversity and independence of each primary algorithm is compared with other algorithms in the crowd after execution. If they have the necessary conditions, they are added to the crowd.

2- Most importantly, in the WOCCE approach, each primary clustering can be inserted into the crowd without affecting other algorithms' results. This approach can detect errors and identify information in the results (by checking the diversity and independence values), and then compensate for these errors with true information from all the results in the crowd, guaranteeing that the errors will not be spread to other members (by changing the total diversity and independence values in each step).

3- In the WOCCE approach, the selection and measurement of independence and diversity are performed in one step. This cause the independence and diversity values to be retained in the final ensemble.

```
Function WOCCE (Dataset, Kb, iT, dT, cT) Return [Result, nCrowd]
nCrowd = 0;
While we have base cluster
    [idx, Basic-Parameter] = Generate-Basic-Algorithm (Dataset, Kb*cT)
    If (Independent (Basic-Parameter) > iT)
        If (Diversity (idx) >dT)
            Insert idx to Crowd-Partitions
            Crowd = Crowd + 1
        End if
    End if
End while
Co-Acc = Make-Correlation-Matrix (Crowd-Partition)
Z = Average-Linkage (Co-Acc)
Result = Cluster (Z, Kb)
```

**Fig. 8.** Pseudo code for the WOCCE framework

Figure (8) shows the pseudo code for the WOCCE procedure. In Figure (8), *Kb* is the number of clusters given by the base algorithm. The *Generate-Basic-Algorithm* function builds the partitions of base clusters (primary results), *Make-Correlation-Matrix* builds the co-association matrix according to equation (9), and the *Linkage* and *Cluster* functions build the final ensemble in accordance with the Average Linkage method. The parameter *Result* is our final ensemble, and *nCrowd* is the number of members in the crowd.

There are two major problems in classic cluster ensemble selection methods; firstly, although the final result is always providing the optimized result in accordance with the selected metrics, there could be *other metrics* to use for generating the best final result. Secondly, it is possible that *all aspects and specifications of data* are not considered or not seen for precise auditing, because in traditional cluster ensemble selection methods, only the basic results are analyzed (including the correct data as well as errors). Thus, it is necessary to pay more attention on other contractive entities in each cluster ensemble algorithm.

Unlike traditional methods, wise clustering uses a structural perspective for generating the best result based on all aspects and specifications of data which operates based on the "The Wisdom of Crowds" theory. The framework of WOCCE includes the four main conditions: Independency of algorithms, diversity of initial (basic) results, decentralization of framework's structure for preserving the quality of final result, and method of feedback combination for safeguarding the auditing results of partitions in the wise crowd (initial results for combination). This structure makes WOCCE a flexible technique and capable of being programed, so that by altering the value of thresholds, It can be adjusted for any data (will be discus in section 4.3).

Furthermore, in WOCCE method all needed information from clustering problems is gathered by *controlling all entities* within cluster ensemble as the result errors in each entity is optimized by information from other entities which consequently reduces the possibility of occurrence of errors in complex data dramatically.

Table (1) presents a brief mapping between terminologies in WOC and the corresponding cluster ensemble area.

**Table 1.** Mapping between WOC and cluster ensemble terminologies

| WOC Terminology | Cluster Ensemble Terminology |
|---|---|
| Primary opinion | Primary partition |
| People | Base or Primary algorithm |
| Wise crowd | Basic or Primary clustering results |
| Diversity of Opinion | Diversity of primary clustering results |
| Opinion independence | Independence of clustering algorithms that generate primary partitions |
| Decentralization | Decentralization in cluster generation |

## 4. Evaluation

This section describes a series of empirical studies and reports their results. In real world, unsupervised methods are used to find meaningful patterns in non-labeled datasets such as web documents. Since the real dataset doesn't have class labels, there is no direct evaluation method for evaluating the performance in unsupervised methods. Like many pervious researches (Fred and Jain, 2005; Fern and Lin, 2008; Alizadeh et al, 2011; Alizadeh et al 2012; Alizadeh et al 2014), this paper compares the performance of its proposed method with other basic and ensemble methods by using standard datasets and their real classes. Although this evaluation cannot guarantee that the proposed method generate high performance for all datasets in comparison with other methods, it can be considered as an example for analyzing the probability of predicting good results in the WOCCE.

### 4.1. Datasets

The proposed method is examined over 14 different standard datasets. We have chosen datasets that are as diverse as possible in their numbers of true classes, features, and samples, as this variety better validates the results obtained. Brief information about these datasets is listed in Table (2). More information is available in Newman et al. (1998) and Alizadeh et al. (2012). The features of the datasets marked with an asterisk are normalized to a mean of 0 and variance of 1, i.e. $N(0,1)$.

**Table 2.** Information about the datasets used in our simulations

| Name | Feature | Class | Sample |
|---|---|---|---|
| **Half Ring** | 2 | 2 | 400 |
| **Iris** | 4 | 3 | 150 |
| **Balance Scale**[*] | 4 | 3 | 625 |
| **Breast Cancer**[*] | 9 | 2 | 683 |
| **Bupa**[*] | 6 | 2 | 345 |
| **Galaxy**[*] | 4 | 7 | 323 |
| **Glass**[*] | 9 | 6 | 214 |
| **Ionosphere**[*] | 34 | 2 | 351 |
| **SA Heart**[*] | 9 | 2 | 462 |
| **Wine**[*] | 13 | 2 | 178 |
| **Yeast**[*] | 8 | 10 | 1484 |
| **Pendigits**[5] | 16 | 10 | 10992 |
| **Statlog** | 36 | 7 | 6435 |
| **Optdigits**[6] | 62 | 10 | 5620 |

### 4.2. Experimental Method for Calculating Thresholds

This paper proposes an experimental method for determining the threshold values *iT, dT,* and *cT*. First, we check the relationships between the thresholds and WOCCE factors:

- *iT* has a relation with the number of base clustering algorithms, the variety of base clustering algorithm types, and the runtime of WOCCE.
- *dT* has a relation with the variety of base clustering algorithm types, the decentralization threshold (*cT*), and the number of partitions in the crowd.
- *cT* has a relation with the number of data in the dataset, the number of features in the dataset, and the number of partitions in the clustering.

In this paper, *cT* is chosen based on the dataset specification such as number of features and samples. Also, all thresholds are chosen such that each WOCCE algorithm's runtime is approximately 30 min on a PC with certain specifications[7].

---

[5] Pen-based recognition of handwritten digits data set
[6] Optical recognition of handwritten digits data set
[7] CPU = Intel X9775 (4*3.2 GHz), RAM = 16 GB, OS = Windows Server 2012 RTM x64

## 4.3. Results

The algorithms in Table (3) were used to generate the wise crowd:

**Table3.** List of base algorithms used in WOCCE

| No. | Algorithm Name |
|---|---|
| 1 | K-Means |
| 2 | Fuzzy C-Means |
| 3 | Median K-Flats |
| 4 | Gaussian Mixture |
| 5 | Subtract Clustering |
| 6 | Single-Linkage Euclidean |
| 7 | Single-Linkage Hamming |
| 8 | Single-Linkage Cosine |
| 9 | Average-Linkage Euclidean |
| 10 | Average-Linkage Hamming |
| 11 | Average-Linkage Cosine |
| 12 | Complete-Linkage Euclidean |
| 13 | Complete-Linkage Hamming |
| 14 | Complete-Linkage Cosine |
| 15 | Ward-Linkage Euclidean |
| 16 | Ward-Linkage Hamming |
| 17 | Ward-Linkage Cosine |
| 18 | Spectral clustering using a sparse similarity matrix |
| 19 | Spectral clustering using Nystrom method with orthogonalization |
| 20 | Spectral clustering using Nystrom method without orthogonalization |

We used MATLAB R2012a (7.14.0.739) in order to generate our experimental results. The distances were measured by a Euclidean metric. All results are reported as the average of 10 independent runs of the algorithm. The final clustering performance was evaluated by re-labeling between obtained clusters and the ground truth labels and then counting the percentage of correctly classified samples. The WOCCE results are compared with well-known base algorithms including Kmeans, Fuzzy Cmeans, Subtract Clustering, and Single-Linkage, as well as five state-of-the-art cluster ensemble methods (EAC, MAX and etc.). Table (4) shows the results.

**Table 4.** Experimental results

| | Primary methods | | | | | | Cluster Ensemble methods | | | | | | |
|---|---|---|---|---|---|---|---|---|---|---|---|---|---|
| | Kmeans | FCM | Subtract | Single Linkage | EAC | MAX | CSPA | HGPA | MCLA | WOCCE | | | |
| | | | | | | | | | | iT | dT | cT | Result |
| *Half Ring* | 75.75 | 78 | 86 | 75.75 | 77.17 | 78.48 | 74.5 | 50 | 74.5 | 0.2 | 0.06 | 3 | **87.2** |
| *Iris* | 65.3 | 82.66 | 55.3 | 68 | **96** | 72.89 | 85.34 | 48.66 | 89.34 | 0.2 | 0.06 | 1 | **96** |
| *Balance Scale* | 40.32 | 44 | 45.32 | 46.4 | 52 | 52.1 | 51.84 | 41.28 | 51.36 | 0.23 | 0.063 | 3 | **54.88** |
| *Breast Cancer* | 93.7 | 94.43 | 65 | 65.15 | 95.02 | 75.72 | 80.97 | 50.37 | 96.05 | 0.18 | 0.02 | 1 | **96.92** |
| *Bupa* | 54.49 | 50.1 | **57.97** | 57.68 | 55.18 | 56.17 | 56.23 | 50.72 | 55.36 | 0.21 | 0.04 | 3 | 57.42 |
| *Galaxy* | 30.03 | 34.98 | 29.72 | 25.07 | 31.95 | 32.78 | 29.41 | 31.27 | 28.48 | 0.2 | 0.05 | 2 | **35.88** |
| *Glass* | 42.05 | 47.19 | 36.44 | 36.44 | 45.93 | 44.17 | 38.78 | 41.12 | 51.4 | 0.19 | 0.06 | 3 | **51.82** |
| *Ionosphere* | 69.51 | 67.8 | **77** | 64.38 | 70.48 | 64.48 | 67.8 | 58.4 | 71.22 | 0.3 | 0.1 | 3 | 70.52 |
| *SA Heart* | 64.51 | 63.41 | 67.26 | 65.15 | 65.19 | 63.96 | 58.42 | 50.93 | 62.54 | 0.65 | 0.8 | 1 | **68.7** |
| *Yeast* | 31.19 | 29.98 | 31.2 | 31.73 | 31.74 | 32.4 | 14 | 15.23 | 17.56 | 0.5 | 0.5 | 1 | **34.76** |
| *Wine* | 65.73 | **71.34** | 67.23 | 37.64 | 70.56 | 69.17 | 67.41 | 62.36 | 70.22 | 0.2 | 0.05 | 3 | **71.34** |
| *Pendigits* | 46.97 | 36.77 | 10.4 | 10.46 | 10.47 | 57.02 | 58.32 | 11.14 | 58.62 | 0.02 | 0.12 | 1 | **58.68** |
| *Optdigit* | 52.52 | 38.33 | 47.72 | 10.28 | 20 | 76.11 | 75.21 | 64.77 | 77.15 | 0.01 | 0.1 | 1 | **77.16** |
| *Statlog* | 50.93 | 49.91 | 23.8 | 23.8 | 23.9 | 54.23 | 54.23 | 52.94 | 55.71 | 0.01 | 0.1 | 1 | **55.77** |

In Table (4) the best results obtained for each dataset have been bolded. As depicted in this table, although basic clustering algorithms have shown high performance in some datasets, they cannot recognize true patterns in all of them. As this paper mentioned, basic algorithms consider an aspect (specification) of a dataset such as density for solving the clustering problem. The results of basic clustering algorithms which are depicted in Table (4) are good evidences for this claim. Furthermore, the CSPA and HGPA results show the effect of the aggregation method on improving accuracy in the final results. According to Table (4), the MCLA, MAX and WOCCE have generated better results in comparison with CSPA and HGPA. Even though WOCCE was outperformed in two datasets (Bupa and Ionosphere) by some algorithms, the majority of results demonstrate superior accuracy for the proposed method. Figure (9) shows the average of accuracy for each technique:

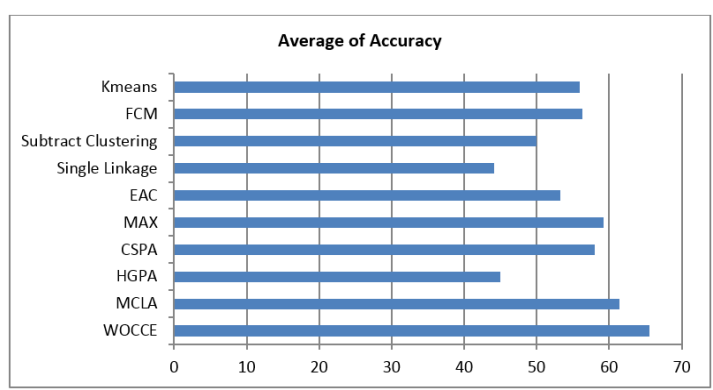

**Fig. 9.** Average of accuracy for each technique

In Figure (9), it is difficult to separate the WOCCE and MCLA methods. However, the average performance over all fourteen datasets reveals that WOCCE outperformed MCLA by over 3%. Also, this figure shows that Single Linkage and EAC generate poor results in comparison with other methods. Although hierarchical methods such as Single Linkage can handle non-elliptical datasets and generate stable results, they are sensitive to noise and outliers particularly in complex datasets (Tan et al, 2005). Needless to say that as a classic ensemble method, there is no evaluation and selection in EAC. This method cannot omit errors which are made in the process of recognizing patterns of the basic clustering results by using the correct information of other basic algorithms' results. The results of EAC which are given in Table (4) and Figure (9) show the effect of evaluation and selection in cluster ensemble selection methods. Since some of the four conditions of the Wisdom of Crowds theory method do not exist in EAC, this method is a good example of unwise crowd. The effect of this method on final results can be seen in Table (4). Table (5) illustrates the NMI rates made by primary and ensemble methods:

**Table5.** Normalized mutual information (NMI) rates

|  | *Primary methods* | | | | | *Cluster Ensemble methods* | | | | |
|---|---|---|---|---|---|---|---|---|---|---|
|  | **Kmeans** | **FCM** | **Subtract** | **Single Linkage** | **EAC** | **MAX** | **CSPA** | **HGPA** | **MCLA** | **WOCCE** |
| *Half Ring* | 0.26 | 0.33 | 0.51 | 0.06 | 0.32 | 0.56 | 0.32 | 0 | 0.34 | **0.68** |
| *Iris* | 0.74 | 0.78 | 0.77 | 0.73 | 0.75 | 0.8 | 0.72 | 0.14 | 0.75 | **0.86** |
| *Balance Scale* | 0.12 | 0.2 | **0.37** | 0.03 | 0.14 | 0.22 | 0.07 | 0.03 | 0.09 | **0.37** |
| *Breast Cancer* | **0.74** | 0.69 | 0.73 | 0.01 | 0.69 | 0.72 | 0.35 | 0 | **0.74** | 0.74 |
| *Bupa* | 0.0008 | 0.0045 | 0 | 0.0136 | 0.0018 | 0.002 | **0.01** | 0 | 0.0016 | 0.0013 |
| *Galaxy* | 0.24 | 0.26 | 0.17 | 0.12 | 0.28 | 0.31 | 0.23 | 0.13 | 0.27 | **0.34** |
| *Glass* | 0.36 | 0.24 | 0.07 | 0.11 | 0.41 | 0.3 | 0.25 | 0.31 | 0.27 | **0.45** |
| *Ionosphere* | 0.12 | 0.08 | 0.07 | 0.02 | 0.11 | 0.12 | 0.1 | 0.02 | 0.13 | **0.15** |
| *SA Heart* | 0.08 | **0.13** | **0.13** | 0 | 0.07 | 0.08 | 0.02 | 0 | 0.08 | 0.078 |
| *Yeast* | 0.1 | 0.11 | 0 | 0.11 | 0.12 | 0.27 | 0.26 | 0.14 | **0.28** | **0.28** |
| *Wine* | 0.75 | 0.55 | 0.72 | 0.05 | 0.69 | 0.79 | 0.77 | 0.43 | 0.8 | **0.83** |
| *Pendigits* | 0.61 | 0.35 | 0 | 0.01 | 0.01 | 0.7 | 0.68 | 0 | 0.69 | **0.71** |
| *Optdigit* | 0.6 | 0.39 | 0.45 | 0.02 | 0.36 | **0.76** | 0.68 | 0.42 | 0.73 | **0.76** |
| *Statlog* | 0.52 | 0.38 | 0 | 0.01 | 0.01 | 0.54 | 0.47 | 0.42 | **0.56** | **0.56** |

In Table (5), the best result obtained for each dataset is highlighted in bold. The NMI evaluation shows the result's stability in each clustering technique (Fred and Jain, 2005; Fern and Lin 2008). As this table illustrates, most of basic clustering algorithms cannot generate robust results in large-scale datasets (see the last three datasets in the table). Even though WOCCE was outperformed in two datasets (Bupa and SA Heart) by some algorithms, the majority of results demonstrate superior NMI for the proposed method.

In the rest of this section, the thresholds' effects on performance and runtime are analyzed. In order to omit the effect of some irrelevant threshold on each experiment, this paper disable this irrelevant threshold by setting it to zero for independence and diversity thresholds and to 1 for decentralization. As mentioned in the previous sections, all proposed metrics in WOCCE can improve the accuracy when they are used at the same time. So, it is obvious that the WOCCE metrics cannot improve the accuracy significantly when other metrics are disabled. In addition, the slope of some line charts has changed slightly, because the real value of that threshold is more than the given values in the experiment. The main goals of our experiment are to show the relation between performance and runtime in WOCCE and to illustrate how this paper determined the optimized values for each threshold.

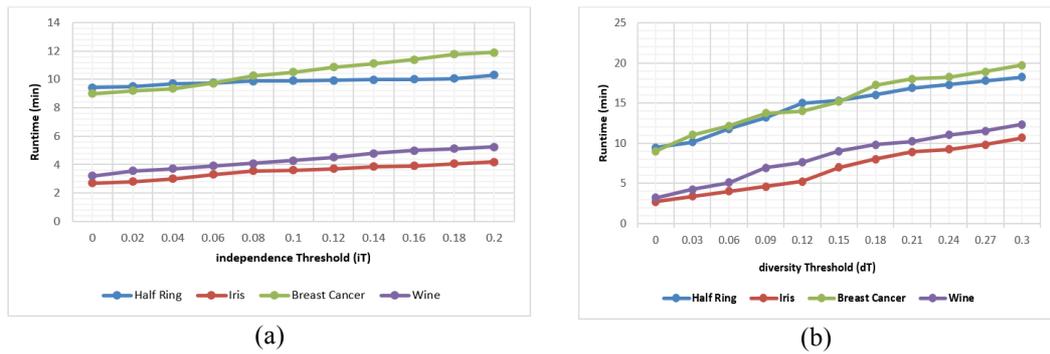

**Fig. 10.** The effect of Independency and Diversity on the runtime

Figure (10) part (a) illustrates the relationship between $iT$ and the runtime of WOCCE. This experiment was performed with $dT = 0$ in order to remove the effect of diversity on the final results. The vertical axis refers to time and the horizontal axis refers to the independence threshold. Figure (10) part (b) illustrates the relationship between $dT$ and the runtime of WOCCE. This experiment was performed with $iT = 0$ in order to remove the effect of independence on the final results. The vertical axis refers to time and the horizontal axis refers to the diversity threshold.

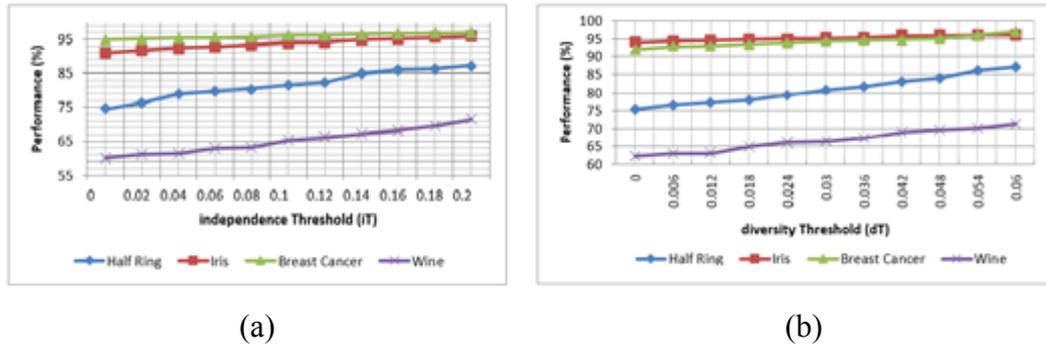

(a)                  (b)

**Fig. 11.** The effect of Independency and Diversity on the WOCCE performance

Figure (11) part (a) and Figure (11) part (b) illustrate the relationship between the performance of WOCCE, which is based on the number of correctly classified samples, and independence and diversity thresholds, respectively. To plot Figure (11) part (a), a fixed value was assigned to *dT* in order to measure the effect of independence in the final results. In Figure (11) part (a), the vertical axis refers to the performance and the horizontal axis refers to the independence. In Figure (11) part (b), *iT* was fixed in order to plot performance with respect to diversity.

Figure (12) illustrates the relationship between the performance of WOCCE, based on the number of correctly classified samples, and the decentralization. To plot Figure (12), the number of clusters in basic clustering results varies between k (original number of dataset classes) to 5×k (cT = 5).

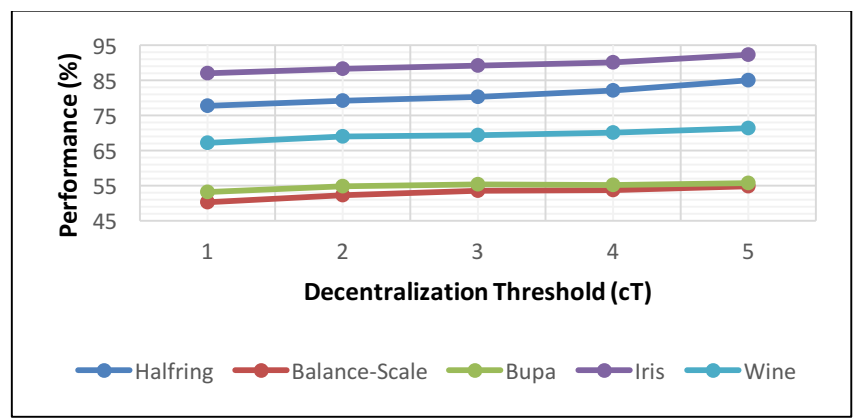

**Fig. 12.** Relation between decentralization and performance

These figures show that the performance increased with the respective threshold value. They illustrate the effect of independence and diversity and decentralization in the performance of our cluster ensemble, and confirm that, along with diversity, the independence is an important factor that should be considered when creating the ensemble.

WOCCE is the first system to date that adds three conditions of diversity in basic results, algorithms' independence, and decentralization in datasets and ensemble components, to the cluster ensemble field. Although we have not presented a mathematical proof to support our method, the experimental results confirm its superior performance with respect to other cluster ensemble methods for most of the benchmark datasets.

## 5. Conclusion

In this paper, the WOC phenomenon was mapped to the cluster ensemble problem. The primary advantage of this mapping was the addition of two new aspects, the independence and decentralization, as well as a new framework to investigate them. Until now, common cluster ensemble methods have concentrated on the diversity of the primary partitions. Inspired by the WOC research in the social sciences, this paper introduced the conditions of independence and decentralization to the field of cluster ensemble research. The proposed WOCCE framework uses a feedback procedure to investigate all three conditions, yielding a wise crowd incorporating decentralization, independence, and diversity.

Similar to other pioneering ideas, the WOCCE framework can be improved later.

This paper suggested employing as different as base algorithm to satisfy the decentralization condition. We also proposed a procedure to assess the independence of the base algorithms, and introduced the A3 criterion to measure the diversity of the partitions. Our suggestions for satisfying the corresponding conditions will be investigated further in future work. The main drawback of the WOCCE algorithm is that it has three threshold parameters that must be set to appropriate values. This parameterization can be considered as another area for future work.